# *Presence-absence reasoning for evolutionary phenotypes*

*James P. Balhoff*[*1,2], *T. Alexander Dececchi*[3], *Paula M. Mabee*[3], *and Hilmar Lapp*[1]

[1]*National Evolutionary Synthesis Center, Durham, NC USA,* [2]*University of North Carolina, Chapel Hill, NC USA,*
[3]*University of South Dakota, Vermillion, SD USA*

## 1 INTRODUCTION

Nearly invariably, phenotypes are reported in the scientific literature in meticulous detail, utilizing the full expressivity of natural language. Both detail and expressivity are usually driven by study-specific research questions. However, research aiming to synthesize or integrate phenotype data across studies or even disciplines is often faced with the need to abstract from detailed observations so as to construct phenotypic concepts that are common across many datasets rather than specific to a few. Yet, observations or facts that would fall under such abstracted concepts are typically not directly asserted by the original authors, usually because they are "obvious" according to common domain knowledge, and thus asserting them would be deemed redundant by anyone with sufficient domain experience. For example, a phenotype describing the length of a manual digit for an organism implicitly means that the organism must have had a hand, and thus a forelimb. In this way, the presence or absence of a forelimb may have supporting data across a far wider range of taxa than the length of a particular manual digit, and may also have wider applications in biological research questions.

For large-scale computational integration of phenotypes the challenge then is, how can machines be enabled to infer such facts that are implied by but not explicitly included in the phenotype observations recorded by the original author(s). As descriptions in natural language, phenotype data require special transformation to become amenable to computational processing to start with. An approach with considerable success in rendering phenotypes computable is to annotate the free text descriptions with ontology terms drawn from anatomy, quality, spatial, taxonomy and other pertinent ontologies, following a common formalism. The aforementioned challenge then is specifically, how can a machine reasoner be enabled to infer implied phenotypes from those asserted, given the anatomy (and other) domain knowledge asserted by ontology axioms in subclass, partonomy, and other hierarchies.

Here we describe how within the Phenoscape project we use a pipeline of axiom generation and inference steps to address this challenge specifically for inferring taxon-specific presence/absence of anatomical entities from anatomical phenotypes. These phenotypes are primarily derived from published comparative anatomical treatments (descriptions of new species or reviews of larger clade interrelationships) in the form of morphological character state matrices, which document for a set of characters the evolutionary patterns of variation (the character states) across a set of taxa (Dahdul et al. 2010). Using the Phenex data annotation tool (Balhoff et al. 2010), Phenoscape curators annotate each character state using the Entity–Quality (EQ) formalism (Mungall et al. 2007, 2010). Anatomical entities are represented by terms from the comprehensive Uberon anatomy ontology for metazoan animals (Haendel et al. 2014), qualities (e.g., presence/absence, size, shape, composition, color, etc.) are drawn from the Phenotype and Trait (PATO) ontology (Gkoutos et al. 2005), and terms for vertebrate taxa are taken from the Vertebrate Taxonomy Ontology (VTO) (Midford et al. 2013). The Phenoscape Knowledgebase (KB, http://phenoscape.org/) is essentially a triple store that integrates such ontology-annotated phenotype data across all studies and data sources and allows querying them.

Although presence/absence is all but one, and a seemingly simple way to abstract phenotypes across data sources, it can nonetheless be powerful for linking genotype to phenotype (Hiller et al. 2012), and it is particularly relevant for constructing synthetic morphological supermatrices for comparative analysis; in fact presence/absence is one of the prevailing character observation types in published character matrices, accounting for 25-50% of data in some large morphological matrices (Sereno 2009).

## 2 OWL REPRESENTATION OF PRESENCE AND ABSENCE

In this section we explain how we represent EQ phenotypes in OWL (Web Ontology Language, http://www.w3.org/TR/2012/REC-owl2-overview-20121211/) so that presence and absence of anatomical structures (the 'E' part) within an organism are reliably inferred, based on asserted knowledge from the anatomy ontology about other structures (here: subclass, partonomy, and developmental relationships).

### 2.1 Presence

Within the Phenoscape KB, a character description annotated with entity 'E' and quality 'Q' is, by default,

translated into an OWL class expression of the form `'Q' and inheres_in some 'E'` (Mungall et al. 2007). Thus each phenotype is a subclass of 'PATO:quality'. The existential restriction entails the existence of an instance of the anatomical entity 'E'. Although strict OWL semantics do not entail that this instance of 'E' exists in the same organism that bears the quality 'Q', common knowledge lets us conclude that 'E' must be present within the organism having this character description. For example, the phenotype `'bifurcated' and inheres_in some 'pectoral fin radial'` implies the presence of a pectoral fin radial, in the organism that bears the phenotype. Some PATO terms are "relational quality" terms, which embody a relation between two structures. E.g., in the phenotype "vertebra is fused with the pelvic girdle", the corresponding PATO term "fused with" is a quality that represents a relation between vertebra and pelvic girdle. The reification of relations as qualities, such as 'fused with', is characteristic of PATO. To deal with such qualities, the Phenex annotation tool provides an optional third entry field besides Entity and Quality, called Related Entity (or 'RE'). For such phenotypes, the OWL expression formed is `'Q' and inheres_in some 'E' and towards some 'RE'`. Hence, the OWL expression for the above example would be `'fused with' and inheres_in some 'vertebra' and towards some 'pelvic girdle'`. Here, too, for the data used within Phenoscape common knowledge says that the instance of 'RE' entailed by OWL semantics must exist in the same organism that bears the phenotype. We therefore introduce an object property *implies_presence_of*, as a super-property of both *inheres_in* and *towards*. For example, the phenotype `'in contact with' and inheres_in some 'internal trochanter' and towards some 'diaphysis of femur'` will be returned in queries using *implies_presence_of* for either 'internal trochanter' or 'diaphysis of femur'.

This model works with the OWL class hierarchy as expected. For example, phenotypes describing the shape of a 'dorsal fin', length of a 'pectoral fin', or color of a 'caudal fin' will all be returned with a query of `implies_presence_of some 'fin'`.

## 2.2 Absence

Using the EQ-to-OWL template described above will provide undesirable results when translating annotations that use the quality 'absent'. For example, for a character description such as "dorsal fin: absent", Phenoscape curators typically annotate: entity = 'dorsal fin'; quality = 'absent'. The default translation would produce the OWL phenotype `'absent' and inheres_in some 'dorsal fin'`. Because of the existential restriction, this expression asserts the existence of a dorsal fin (Hoehndorf et al. 2007, Mungall et al. 2010), even though the observation means to state that there is no such instance in the respective organism. Additionally, reasoning with such classes produces unintuitive results. An organism with no fins at all would according to the above template have the phenotype `'absent' and inheres_in some 'fin'`. An OWL reasoner would, correctly, infer from this that the absence of a dorsal fin is a subclass of the absence of fins, which is the opposite of what we really intend—organisms without dorsal fins should be a superset of the organisms without any fins. That is, every organism with no fins necessarily does not have a dorsal fin. However, there are organisms with no dorsal fin that do have other fins. These expressions also do not provide a means to delineate whether the structure is absent from the whole organism or instead just from one part (e.g. feathers absent from head).

The solution provided by PATO for problems with the 'absent' quality is the relational quality 'lacks all parts of type' (Mungall et al. 2010). Instead of describing an absence which inheres in a dorsal fin, we can instead describe the lack of dorsal fin which inheres in the whole body. An expression using an existential restriction on *towards*—`'lacks all parts of type' and inheres_in some 'body' and towards some 'dorsal fin'`—produces the same classification problem as before for absent "fins" and "dorsal fins". But because 'lacks all parts of type' is referring not to a particular instance, but instead to the whole class 'dorsal fin', we would like to refer directly to the class itself within the expression (Hoehndorf et al. 2007). Within OWL DL, we can use "punning" to simulate reference to the class by using an OWL individual with the same identifier (as the class) as the value for the *towards* relation: `'lacks all parts of type' and inheres_in some 'body' and towards value 'dorsal fin'`. While this expression seems to capture the intended absence, and does prevent the unintended reasoning problems described above, using a class identifier as an instance value will also prevent OWL reasoners from making any useful inferences with respect to the class hierarchy of the absent structures.

Fortunately, we can explicitly provide semantics for these expressions, by asserting that, for every entity 'E', `'lacks all parts of type' and towards value 'E'` is equivalent to `inheres_in some (not (has_part some 'E'))`. By standard OWL semantics, given classes 'A' and 'B' and axiom `B SubClassOf A`, the complement of 'A'—(not 'A')—will be a subclass of the complement of 'B'. Thus, when expressed in this way, our system will correctly treat the absence of fins as a subclass of the absence of dorsal fins.

Within Phenoscape, we keep the expression involving 'lacks all parts of type', along with the additional "not has part" semantics, since it retains a parallel structure to our other, non-absence, phenotype expressions. Also, PATO provides related terms, such as 'has fewer parts of type', which allow annotation of concepts for which the full semantics cannot be directly expressed within OWL DL (Mungall et al. 2010).

## 2.3 Extending inference of presence and absence

While this presence/absence model works correctly across the basic OWL class hierarchy for anatomical structures, we would like to leverage the knowledge encoded within the ontology to make further inferences. Specifically, we would like to infer presence and absence across partonomic and developmental existential relations. For this we leverage the assumption that for the phenotype data we collect in the KB anatomical structures are only part of, and only have parts, that are part of the same organism (i.e., organisms are never asserted to be members of some larger grouping via *part_of*). Thus we provide the following property chains for *implies_presence_of*:

`implies_presence_of ∘ part_of → implies_presence_of`
`implies_presence_of ∘ has_part → implies_presence_of`

These property chains entail that for any phenotype that implies the presence of an entity E, also implied is the presence of all entities E′ for which the ontology contains axioms E `subClassOf` (`part_of` some E′), E `subClassOf` (`has_part` some E′), respectively. For example, if the ontology (Uberon in this case) asserts that every 'humerus' is part of some 'forelimb', then a phenotype that implies the presence of 'humerus' also implies the presence of a 'forelimb' in that organism. Similarly we would like to infer that an organism has (or at least had at some point during its development) any structure that one of its known to be present structures develops from:

`implies_presence_of ∘ develops_from →`
`implies_presence_of`

Thus we can infer that, when using the Uberon anatomy ontology, any vertebrate animal that has a limb must have had a limb bud, since `'limb' SubClassOf develops_from some 'limb bud'`. Although the definition of *develops_from* implies that the presence of limb bud and limb were distinct in time during the individual's existence (Smith et al. 2005), our applications of these inferences are agnostic to when an entity was present during an individual organism's lifetime. If time of presence is important, inferring presence from developmental relationship will require an explicit temporal context.

To fully extend the knowledge captured in the anatomy ontology, absence must also propagate correctly over *develops_from*, *has_part*, and *part_of*. For absence we obtain inferences that are the inverse of the presence entailments. For example, with the above axiom of all limbs developing from some limb bud, if it is asserted that an organism has no limb buds (at any time during its development), we should be able to infer that it must also lack limbs. This requires the addition of another property chain:

`has_part ∘ develops_from → has_part`

For a given entity, such as 'limb bud', this implies that `has_part` some (`develops_from` some 'limb bud') is a subclass of `has_part` some 'limb bud'. The negations of these classes then have the reverse subclass relationship: `not` (`has_part` some 'limb bud') is a subclass of `not` (`has_part` some (`develops_from` some 'limb bud')). So organisms that have absent limb buds can now be inferred to lack anything asserted within the anatomy ontology to develop from a limb bud.

We would like the same reasoning to apply across *has_part* and *part_of*. For example, any organism which lacks a structure, e.g. 'forelimb', must also lack any structures asserted to be part of it, e.g. 'humerus'. The inverse property axioms between the *has_part* and *part_of* properties prevent us from achieving the desired inference by asserting a property chain such as the following:

`has_part ∘ part_of → has_part`

This would result in circular dependencies between both properties, which is forbidden in OWL 2 DL. As a workaround, for every anatomical structure 'E', we generate the following axiom:

(`has_part` some (`part_of` some 'E')) SubClassOf (`has_part` some 'E')

This yields the desired result. For example, for 'forelimb' and 'humerus' the generated axiom `not` (`has_part` some 'forelimb') is inferred to be a subclass of `not` (`has_part` some (`part_of` some 'forelimb')), and thus also of `not` (`has_part` some 'humerus'). This axiom generation is fully automated within the Phenoscape KB build tools.

## 3 SCALING PRESENCE/ABSENCE REASONING

The approach described in section 2 works with any complete OWL DL reasoner such as HermiT (Shearer et al. 2008), and is implemented in a demonstration ontology (http://purl.org/phenoscape/demo/presence_absence.owl).

However, in our experience no OWL DL reasoner scales adequately to handle a single annotated morphological character matrix dataset. In fact, we have not found any OWL DL reasoner that can classify the sizable Uberon anatomy ontology, even without the introduction of any phenotype annotation data. Thus, in production we are constrained to use the highly scalable ELK reasoner (Kazakov et al. 2013) for all OWL reasoning tasks. Because ELK implements only the OWL EL profile, it does not support the use of inverse properties or, more importantly for absence reasoning, class negation. For this reason we have implemented several workarounds within the Phenoscape KB build tools which provide the needed inferences in conjunction with ELK.

To support phenotype queries within the KB application using absence expressions, we must assert the complete class hierarchy of anatomical absences in advance, since ELK cannot classify them on its own. However, we can use ELK to help generate this hierarchy. In addition to generating the axioms described in the previous section, the Phenoscape KB build system performs the following steps:

(1) For every anatomical structure, generate a named class for its absence, using the OWL representation described above. For example:

    <absent_dorsal_fin> EquivalentTo 'lacks all parts of type' and *towards* value 'dorsal fin'
    <absent_dorsal_fin> EquivalentTo *inheres_in* some (not (has_part some 'dorsal fin'))

(2) For every anatomical structure, generate a named class as equivalent to has_part some 'E', and another as equivalent to not (has_part some 'E').

(3) Relate each named "not has part" class to its named complement, using an annotation property ("*negates*").

(4) Classify the entire dataset using ELK, and materialize all non-redundant inferred subclass axioms. ELK will not generate any classification for the "not has part" classes.

(5) Create a classification for the "not has part" classes by processing each of the class pairs related via the *negates* annotation. For A *negates* B, each of the direct superclasses of 'B' will be asserted to be subclasses of 'A', and each of the direct subclasses of 'B' will be asserted to be superclasses of 'A'.

(6) Reclassify the dataset using ELK, which will now have enough information to adequately compute the class hierarchy of absences.

This workflow is implemented within the Phenoscape KB build system, available on GitHub at https://github.com/phenoscape/phenoscape-owl-tools.

## 4 APPLICATION

To demonstrate the potential of the described presence-absence inference workflow, we applied our method to a published morphological character matrix (Ruta 2011) that has been annotated by the Phenoscape project with EQ phenotype expressions (Mungall et al. 2010) as described earlier (Dahdul et al. 2010). The matrix describes the appendicular morphology of 43 lobe-finned fish and early tetrapods and consists of 157 descriptive characters. Our workflow generates from this source matrix a new character matrix of asserted and inferred presence/absence knowledge for any subclasses of UBERON:'anatomical structure', resulting in 938 "characters" (i.e. anatomical classes that may be present or absent in a taxon). 872 of these had no direct assertions of presence or absence in the source matrix, but their presence or absence for at least some taxa is inferred by our method. About half (19,222 of 40,334) of the matrix cells are populated, of which only 8% (1451) result from direct presence or absence assertions in the source matrix. Hence, 92% of the populated cells are the result of inference. This single source-matrix workflow is available in executable form that can be repeated for the example matrix described, or applied to other matrices (see http://dx.doi.org/10.5281/zenodo.10071).

The example shows that data inferred through our workflow can substantially supplement those directly asserted, and other matrices we have annotated yield similar results. We are currently developing a tool that utilizes the Phenoscape KB to generate presence/absence supermatrices, to combine information across multiple studies for an anatomical and taxonomic slice chosen by the user.

## ACKNOWLEDGMENTS

We thank Chris Mungall for discussions of representing absence in OWL. The Phenoscape project is funded by NSF (DBI-1062404 and DBI-1062542), and supported by the National Evolutionary Synthesis Center (NESCent), NSF EF-0423641.